\documentclass[letterpaper]{article} 
\usepackage{aaai2027}  
\usepackage[hyphens]{url}  
\usepackage{graphicx} 
\usepackage{natbib}  
\usepackage{caption} 
\usepackage{algorithm}
\usepackage{algorithmic}
\usepackage{amsmath}
\usepackage{amssymb}
\usepackage{booktabs}
\usepackage{multirow}
\usepackage[table]{xcolor}
\definecolor{ourrow}{gray}{0.90}
\definecolor{grouprow}{gray}{0.94}
\definecolor{trajcol}{gray}{0.93}
\newcommand{\yes}{\checkmark}
\newcommand{\partialyes}{\textcolor{black!55}{$\circ$}}
\newcommand{\no}{\textcolor{black!30}{$\times$}}
\newcommand{\na}{\textcolor{black!30}{---}}
\newcommand{\method}{\textsc{PGS}}
\newcommand{\bench}{\textsc{DynaHall}}

\newcommand{\ourAccGain}{3.0}       

\title{Before the Token Commits: Trajectory-Level Benchmarking \\ of Visual Hallucinations in Diffusion VLMs}

\author{
Yadong Wang\textsuperscript{\rm 1,2},
Siping Yue\textsuperscript{\rm 1},
Yu Tian\textsuperscript{\rm 3},
Chuanxing Geng\textsuperscript{\rm 1},
Xiang Chen\textsuperscript{\rm 1,2}\corresponding
}

\affiliations{
\textsuperscript{\rm 1}Nanjing University of Aeronautics and Astronautics\\
\textsuperscript{\rm 2}State Key Laboratory of Ocean Sensing,
ZJU-Hangzhou Global Scientific and Technological Innovation Center,
Zhejiang University, Hangzhou, 311215, China\\
\textsuperscript{\rm 3}360 Security Capability Center\\
adamwang373@nuaa.edu.cn, xiang\_chen@nuaa.edu.cn
}

\begin{document}
\maketitle

\begin{abstract}
Multimodal diffusion language models generate responses by iteratively unmasking tokens, making each answer the endpoint of a multi-step trajectory rather than an immediate commitment. Hallucination benchmarks built for autoregressive models evaluate only the final output, and therefore cannot determine whether an unsupported claim in diffusion VLMs appears late or has already stabilized before any answer token is revealed. We introduce \textbf{\bench{}}, a trajectory-level benchmark of annotation-backed binary visual propositions covering object existence, counting, attributes, and relations, with controlled hard negatives graded by visual prior. \bench{} is paired with a commitment-aware protocol that records the intermediate answer tendency at every unmasking step alongside the committed output. Across five diffusion VLMs from three architecture families, visual hallucination is settled before commitment: an unsupported answer is already the preferred state while the answer position is still masked, and later unmasking steps rarely reverse it, so the failure is not introduced at the write step. This holds across decoding schedules, answer formats, and open-ended generation. \bench{} also exposes failures hidden by final-output metrics, including counting and relation collapse, prior-driven false positives, and attribute errors whose direction changes by type. Guided by this diagnosis, \textbf{\method{}} (\textbf{P}re-commitment \textbf{G}radient \textbf{S}teering) edits still-masked answer states to reduce false positives, bringing the affirmation rate close to balance, and transfers to another architecture without degrading general ability. \bench{} and \method{} suggest that hallucination should be measured and mitigated along the generation trajectory of diffusion VLMs, not only at the final answer.
\end{abstract}

\section{Introduction}

Vision-language models (VLMs) are now widely used for visual instruction following, image understanding, and multimodal reasoning~\citep{DBLP:conf/nips/Dai0LTZW0FH23,DBLP:journals/corr/abs-2502-13923,DBLP:conf/cvpr/LiuLLL24}. Recent multimodal diffusion language models depart from the dominant autoregressive generation paradigm by refining masked responses through iterative unmasking, rather than generating tokens strictly from left to right~\citep{DBLP:conf/nips/AustinJHTB21,DBLP:journals/corr/abs-2505-16933,DBLP:conf/nips/YangTLZSTW25,DBLP:journals/corr/abs-2505-16839,DBLP:conf/icml/LouME24}. In this paper, we study these models in the visual-understanding setting and refer to them as diffusion VLMs. Despite this progress, reliable visual grounding remains difficult. Models may still assert objects, attributes, counts, or relations that are not supported by the image, a failure mode commonly known as visual hallucination. Existing hallucination benchmarks have made this problem measurable through caption-level object metrics, yes/no visual probing, multidimensional visual propositions, and expert-designed diagnostic cases. However, many widely used benchmarks evaluate hallucination only after generation is complete, by comparing a final answer, caption, or yes/no response with image evidence.

\begin{figure}[t]
\centering
\includegraphics[width=\columnwidth]{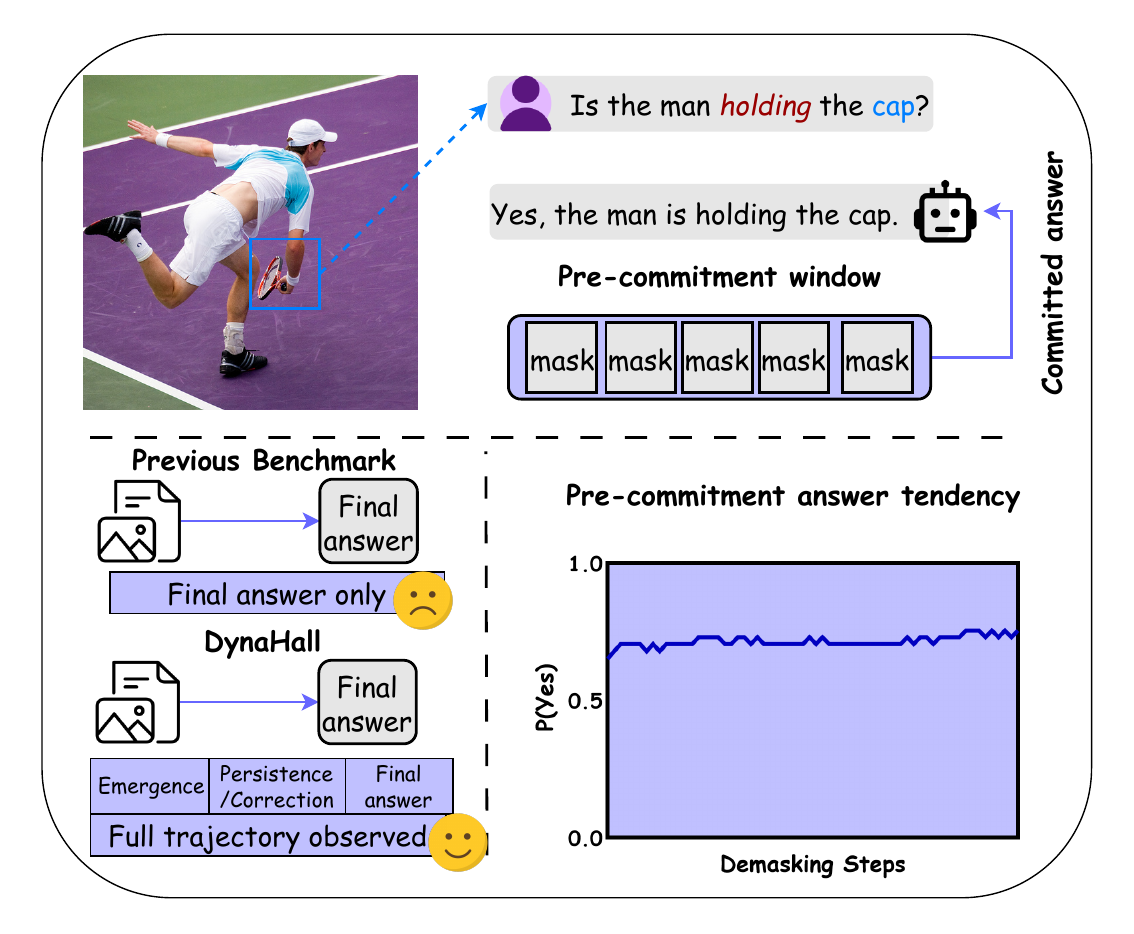}
\caption{Final-output benchmarks observe only the committed endpoint, whereas DynaHall tracks the full unmasking trajectory. The example shows a hallucinated answer tendency that is already stable before the answer token is committed.
}
\label{fig:teaser}
\end{figure}

\begin{table*}[t]
\centering
\footnotesize
\setlength{\tabcolsep}{4pt}

\begin{tabular*}{\textwidth}{@{\extracolsep{\fill}}lccccccc@{}}
\toprule
 & \multicolumn{2}{c}{Target} & \multicolumn{4}{c}{Evaluation design} & Method \\
\cmidrule(lr){2-3}\cmidrule(lr){4-7}\cmidrule(lr){8-8}
Benchmark / study & Visual & Diffusion & Trajectory & Graded neg.\ & Bias-ctrl.\ & Annot.\ & Mitigation \\
\midrule
\rowcolor{grouprow}\multicolumn{8}{@{}l}{\textit{LVLM hallucination benchmarks (autoregressive)}}\\
POPE~\cite{DBLP:conf/emnlp/LiDZWZW23}            & \yes & \no & \no & \yes & \yes & \yes & \no \\
AMBER~\cite{DBLP:journals/corr/abs-2311-07397}           & \yes & \no & \no & \no & \partialyes & \yes & \no \\
HallusionBench~\cite{DBLP:conf/cvpr/GuanLWXLL0CHYM024}  & \yes & \no & \no & \partialyes & \yes & \no & \no \\
Reefknot~\cite{DBLP:conf/acl/ZhengCY00H25}        & \yes & \no & \no & \partialyes & \partialyes & \yes & \yes \\
\midrule
\rowcolor{grouprow}\multicolumn{8}{@{}l}{\textit{Diffusion-LM hallucination (text-only)}}\\
TraceDet~\cite{DBLP:journals/corr/abs-2510-01274}            & \no & \yes & \yes & \na & \na & \na & \no \\
TDGNet/DynHD/HIVE$^{\dagger}$   & \no & \yes & \yes & \na & \na & \na & \no \\
\midrule
\rowcolor{grouprow}\multicolumn{8}{@{}l}{\textit{Multimodal diffusion hallucination (mitigation)}}\\
VISAGE~\cite{DBLP:journals/corr/abs-2603-25711}              & \yes & \yes & \no & \na & \na & \na & \yes \\
\midrule
\rowcolor{ourrow}
\textbf{\bench{} (ours)} & \yes & \yes & \yes & \yes & \yes & \yes & \yes \\
\bottomrule
\end{tabular*}
\caption{Positioning of \bench{} against representative hallucination benchmarks and diffusion-LM trajectory studies. \checkmark/$\circ$/$\times$ denote yes/partial/no; $^{\dagger}$~\cite{DBLP:journals/corr/abs-2602-08048,DBLP:journals/corr/abs-2603-16459,DBLP:journals/corr/abs-2604-26139}.}
\label{tab:positioning}
\end{table*}

This final-output view usefully measures whether a completed response is faithful, but leaves an important question unanswered for diffusion VLMs. Unlike autoregressive VLMs, which commit tokens sequentially, diffusion VLMs maintain a partially masked response, updating it through iterative unmasking under bidirectional context~\cite{DBLP:journals/corr/abs-2505-16933}. During this process, an answer position may remain uncommitted even when the model already provisionally prefers a candidate answer. As illustrated in Figure~\ref{fig:teaser}, a final false positive may follow different trajectories that final-output evaluation cannot separate, obscuring when unsupported visual claims are settled and whether they persist before being written into the output. Table~\ref{tab:positioning} situates \bench{} against prior hallucination benchmarks and diffusion-trajectory studies along these axes.

We introduce \textbf{\bench{}} (\emph{\textbf{Dyna}mic \textbf{Hall}ucination}), a trajectory-level benchmark for visual hallucination in diffusion VLMs. \bench{} contains $8{,}000$ annotation-backed binary visual propositions covering object existence, counting, visual attributes, and relations. Each example is paired with controlled hard negatives, including plausible absent objects, off-by-one counts, attribute swaps, and predicate swaps, so that success requires grounding the queried proposition in the image rather than relying on language priors. Beyond final yes/no correctness, \bench{} records the unmasking trajectory at the answer locus. At each step, it tracks the provisional answer tendency of the model while the answer position remains masked, together with the committed output state after the answer token is revealed. This commitment-aware protocol supports standard final-output metrics while enabling trajectory-level measurements of emergence, persistence, correction, drift, and answer flips.

\begin{figure*}[t]
\centering
\includegraphics[width=\textwidth]{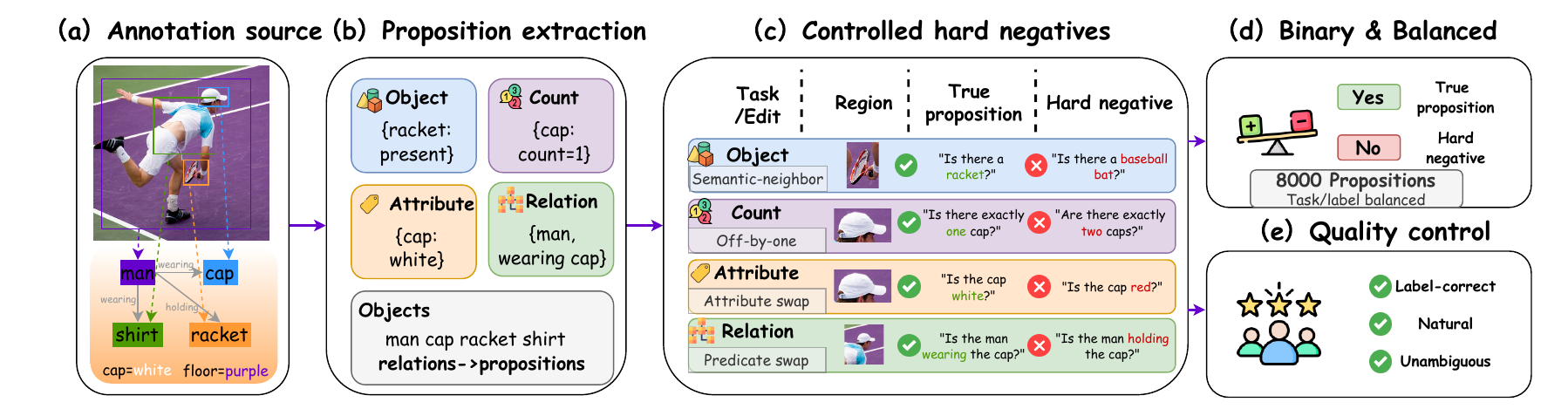}
\caption{Construction of \bench{}. From COCO instances and GQA scene graphs (a), we build binary visual propositions for four tasks (b), each paired with a controlled, prior-graded hard negative (c); both polarities form a balanced set of $8{,}000$ propositions (d) validated by expert audit (e). A single tennis scene runs through every stage.}
\label{fig:benchmark}
\end{figure*}

We evaluate five diffusion VLMs across three architecture families. Early answer convergence has been reported for correct answers in text diffusion LMs~\cite{DBLP:journals/corr/abs-2508-19982}; we find that it holds equally for answers the image does not support: by the time the answer is written, an unsupported tendency is already fixed, even under adversarial prior pressure where later evidence integration might have corrected it. Across the models we study, Figure~\ref{fig:flatness} shows that stage-wise false-positive rates change only weakly over much of the unmasking process, suggesting that unsupported tendencies often appear before the final answer token is revealed rather than only as final-step accidents. \bench{} also exposes failure modes that aggregate final accuracy can hide. Counting and relation propositions are substantially harder than object-existence propositions, plausible hard negatives amplify prior-driven false positives, and low false-positive rates can sometimes reflect conservative under-answering rather than stronger visual grounding. Visual attributes are not uniform either: shape queries tend to trigger over-affirmation, whereas material queries more often trigger conservative denial. The counting and relation collapse, moreover, is not specific to diffusion decoding (a strong autoregressive VLM exhibits the same task profile~\cite{DBLP:journals/corr/abs-2502-13923}), indicating that \bench{} surfaces failure modes shared across generation paradigms rather than artifacts of one decoder.

Although \bench{}'s contribution is diagnostic and does not depend on any particular fix, the diagnosis naturally raises whether the pre-commitment signal it exposes can be used for intervention. We test this with \textbf{\method{}}, a lightweight, diagnosis-driven intervention that steers still-masked answer states during unmasking. On \bench{}, \method{} reduces false positives and improves accuracy while moving the affirmation rate toward balance, and it transfers to a second diffusion VLM without degrading general ability on external benchmarks.

Our contributions are summarized as follows:
\begin{itemize}
\item We introduce \bench{}, a trajectory-level hallucination benchmark for diffusion VLMs: $8{,}000$ annotation-backed propositions with controlled hard negatives across object existence, counting, attributes, and relations.

\item We propose a commitment-aware protocol that separates provisional answer tendencies from committed tokens, enabling trajectory metrics for emergence, persistence, correction, drift, and answer flips.

\item Across five diffusion VLMs, we show that many false positives are decided before the answer token is written, with failures shaped by task, hard-negative pressure, and answer bias. Guided by this diagnosis, \method{} edits still-masked states to reduce these errors.
\end{itemize}

\section{The \bench{} Benchmark}

\bench{} evaluates visual hallucination in diffusion VLMs along the unmasking trajectory rather than only at the final output, which calls for a benchmark structure different from standard captioning or VQA evaluation. Three requirements follow: the queried visual fact must be assessable at every intermediate step, the negatives must test visual grounding rather than shallow priors, and provisional answer preferences must be separable from committed output tokens. \bench{} meets these with binary visual propositions parsable into \{`yes', `no', unresolved, invalid\} throughout unmasking, controlled hard negatives built by minimal edits, and a commitment-aware protocol that records both intermediate tendencies and token commitments. The resulting construction pipeline is shown in Figure~\ref{fig:benchmark}.

\subsection{Benchmark Construction}
\paragraph{Visual propositions.} Each example pairs an image with a binary question, a gold label, a structured proposition, and annotation-backed evidence. \bench{} contains $8{,}000$ examples across four tasks built from COCO val2017~\cite{DBLP:conf/eccv/LinMBHPRDZ14} instance annotations (object existence, counting) and GQA~\cite{DBLP:conf/cvpr/HudsonM19} validation scene graphs (attributes, relations), ensuring every label is annotation-backed rather than LLM-generated. The benchmark is balanced by construction: each task contributes $1{,}000$ positive and $1{,}000$ negative propositions, split into a $1{,}000$-example development set and a $7{,}000$-example test set over disjoint images while approximately preserving task and answer balance. Table~\ref{tab:dataset} summarizes the task and label balance.

\begin{table}[t]
\centering
\footnotesize
\setlength{\tabcolsep}{7pt}

\begin{tabular*}{\columnwidth}{@{\extracolsep{\fill}}lrrr@{}}
\toprule
Task & \#Pos & \#Neg & Total \\
\midrule
Object existence & 1{,}000 & 1{,}000 & 2{,}000 \\
Counting         & 1{,}000 & 1{,}000 & 2{,}000 \\
Attribute        & 1{,}000 & 1{,}000 & 2{,}000 \\
Relation         & 1{,}000 & 1{,}000 & 2{,}000 \\
\rowcolor{ourrow}\textbf{Total} & \textbf{4{,}000} & \textbf{4{,}000} & \textbf{8{,}000} \\
\midrule
\rowcolor{grouprow}\multicolumn{4}{@{}l}{\textit{Construction summary}}\\
\multicolumn{3}{@{}l}{Positive\,:\,negative ratio} & 1\,:\,1 \\
\multicolumn{3}{@{}l}{Hard-negative subtypes} & 6 \\
\multicolumn{3}{@{}l}{Image sources (COCO\,/\,GQA)} & 4{,}000\,/\,4{,}000 \\
\multicolumn{3}{@{}l}{Unique images} & 6{,}395 \\
\bottomrule
\end{tabular*}
\caption{Dataset statistics for \bench{}. The benchmark is balanced by task and answer label.}
\label{tab:dataset}
\end{table}

\paragraph{Controlled hard negatives.} Negatives are minimally altered false variants of plausible claims, graded by language-prior pressure. Object existence uses a three-level ladder consisting of random, co-occurring, and semantic-neighbor absent objects; counting perturbs the true count by one; attributes replace an annotated property with an in-group alternative (color, material, shape); relations keep the object pair fixed and swap the predicate. These controlled negatives test whether a model grounds the proposition in the image rather than relying on language priors or co-occurrence patterns.

\paragraph{Quality control.} We apply automatic checks for task and answer balance, image-disjoint splits, de-duplication, per-image caps, negative-label consistency, and filters for size, crowding, and ambiguity. Three graduate-student annotators then verify a stratified sample of $820$ items covering every task, label, and negative subtype: $97.9\%$ of labels are confirmed correct ($96.5\%$ on counting, $98.5\%$ on relations), with only $0.5\%$ flagged ambiguous. Measured label noise is thus an order of magnitude below the $62$--$64\%$ error rates the hardest tasks elicit, so those rates reflect model failures rather than annotation artifacts.

\subsection{Trajectory Evaluation}
Unlike final-output benchmarks, \bench{} reads each answer along the full unmasking trajectory as the response is gradually formed at the answer locus, rather than scoring only the final committed response. The resulting metrics support the pre-commitment diagnosis central to our analysis and cannot be computed from final outputs alone.

\begin{figure}[t]
\centering
\includegraphics[width=\columnwidth]{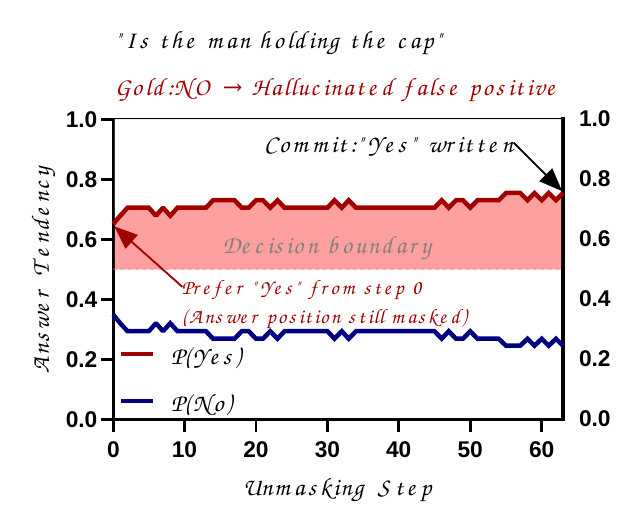}
\caption{Trajectory readout for one running example (``Is the man holding the cap?'', gold no): the answer tendency P(yes)/P(no) at the locus $p$ across unmasking steps, with the answer token committed only at the final step.}
\label{fig:trajectory}
\end{figure}

\begin{table*}[t]
\centering

\footnotesize
\setlength{\tabcolsep}{4pt}
\begin{tabular*}{\textwidth}{@{\extracolsep{\fill}}llrrrr*{6}{>{\columncolor{trajcol}}r}@{}}
\toprule
 & & \multicolumn{4}{c}{Final output} & \multicolumn{6}{>{\columncolor{trajcol}}c}{\textit{Trajectory dynamics (\bench{})}} \\
\cmidrule(lr){3-6}\cmidrule(lr){7-12}
Model & Family & Acc$\uparrow$ & FP$\downarrow$ & FN$\downarrow$ & Aff.\ & Early-FP & Emerg.\ & Persist.\ & Correct.\ & Drift & Flips \\
\midrule
MMaDA-Base   & MMaDA/VQ   & 66.2 & 42.5 & 25.0 & 58.7 & 96.5 & 0.14 & 28.0 & 21.8 & 9.0 & 1.08 \\
MMaDA-MixCOT & MMaDA/VQ   & 74.1 & 14.5 & 37.5 & 38.4 & 78.8 & 0.15 & 20.9 & 19.9 & 6.7 & 0.99 \\
Lumina-DiMOO & VQ-diff.   & 73.6 & \textbf{6.4} & 46.6 & 29.8 & 92.0 & 0.06 & 24.7 & 8.6 & 3.5 & 0.23 \\
LLaDA-V$^{*}$ & cont.-img & 75.9 & 19.7 & \textbf{13.8} & 46.8 & 76.9 & 0.08 & 15.9 & 27.7 & 12.0 & 0.50 \\
Dimple-7B & Qwen-diff. & \textbf{81.4} & 19.8 & 17.5 & 51.0 & 91.8 & 0.14 & 16.3 & 22.7 & 4.7 & 0.19 \\
\bottomrule
\end{tabular*}
\caption{Model-level hallucination diagnosis on the \bench{} test set (percentages). Final-output columns are standard metrics; the shaded Trajectory-dynamics columns are unique to \bench{} and describe how errors evolve, with no preferred direction (Emergence is a normalized step, Flips a count). Best per oriented column ($\uparrow$/$\downarrow$) in \textbf{bold}. Correction/Drift are the shares of ever-wrong/ever-correct trajectories that recover or later flip; Persistence is the share of all trajectories that remain wrong continuously from their first erroneous tendency through the final output. $^{*}$LLaDA-V's $7.3\%$ invalid outputs count as accuracy errors and as non-affirmative for Aff.; they enter neither the FP nor FN numerator.}
\label{tab:dynahall_leaderboard}
\end{table*}

\paragraph{Answer locus and states.} Figure~\ref{fig:trajectory} illustrates how \bench{} records answer states throughout iterative unmasking. Let $p$ denote the answer locus, namely the response position where the binary answer is emitted. At each step, we record two states. The tendency state represents the preferred answer of the model at $p$. It is read from the current logits and parsed into \{`yes', `no', unresolved, invalid\}, capturing the answer favored by the model before commitment. The commitment state is parsed only from unmasked tokens and remains unresolved while $p$ is masked, capturing what has actually entered the response. Together, the two states separate when the model starts to favor a hallucinated claim from when that claim is written. This distinction matters because preference can emerge before the corresponding token is unmasked. To compare diffusion VLMs, we follow the native inference path of each model and normalize trajectory time to $[0,1]$. For single-word answerers, the answer locus is the first response position. For conversational models whose responses may begin with explanatory text, we anchor the answer locus at the first generated `yes' or `no' token to avoid treating a reasoning token as the answer. We report stage-wise statistics at masked ratios $\{0.8,0.6,0.4,0.2,\text{final}\}$ and retain the full per-step trajectory.

\paragraph{Metrics.} Alongside standard final-output metrics (accuracy, FP/FN rates, precision/recall/F1, the invalid rate, and the \emph{affirmation rate}, which makes answer bias explicit), \bench{} reports \emph{trajectory} metrics computable only from its commitment-aware readout: \textbf{Early-FP} (the fraction of final false positives already favoring the unsupported answer at $80\%$ masked), \textbf{Emergence} (the normalized step at which the final answer stabilizes; larger is later), \textbf{Persistence} (the share of all trajectories that remain wrong continuously from their first erroneous tendency to the final output), \textbf{Correction} (the share of ever-wrong trajectories that recover), \textbf{Drift} (the share of ever-correct trajectories that end wrong), and \textbf{Flips} (the number of yes/no reversals along the trajectory). 

\section{Diagnosing Diffusion VLMs}
\subsection{Experimental Setup}
We evaluate five diffusion VLMs from three architecture families on the full \bench{} test set: MMaDA-8B-Base~\cite{DBLP:conf/nips/YangTLZSTW25}, MMaDA-8B-MixCOT, Lumina-DiMOO~\cite{DBLP:journals/corr/abs-2510-06308}, LLaDA-V~\cite{DBLP:journals/corr/abs-2505-16933}, and Dimple-7B~\cite{DBLP:journals/corr/abs-2505-16990}. Only the MMaDA pair shares a backbone, so patterns across all five point to diffusion generation rather than architecture-specific artifacts. Each model uses its official inference path under both protocols with greedy decoding, yielding deterministic paired comparisons; we report bootstrap confidence intervals and McNemar's test for paired significance, with per-model details in the supplement.

\subsection{Pre-commitment of Visual Hallucination}
Table~\ref{tab:dynahall_leaderboard} summarizes the results for the five models. Final accuracy ranges from $66.2\%$ for MMaDA-8B-Base to $81.4\%$ for Dimple-7B, with the high-affirmation MMaDA-Base showing the highest hallucination rate. A low false-positive rate, however, does not necessarily indicate stronger grounding. Lumina-DiMOO has the lowest FP rate, but also the lowest affirmation rate and the highest FN rate, indicating a conservative response pattern. Because the evaluated models span three architecture families, we can further examine whether the timing of hallucination remains consistent even when its magnitude differs across models.

\begin{figure}[t]
\centering
\includegraphics[width=\columnwidth]{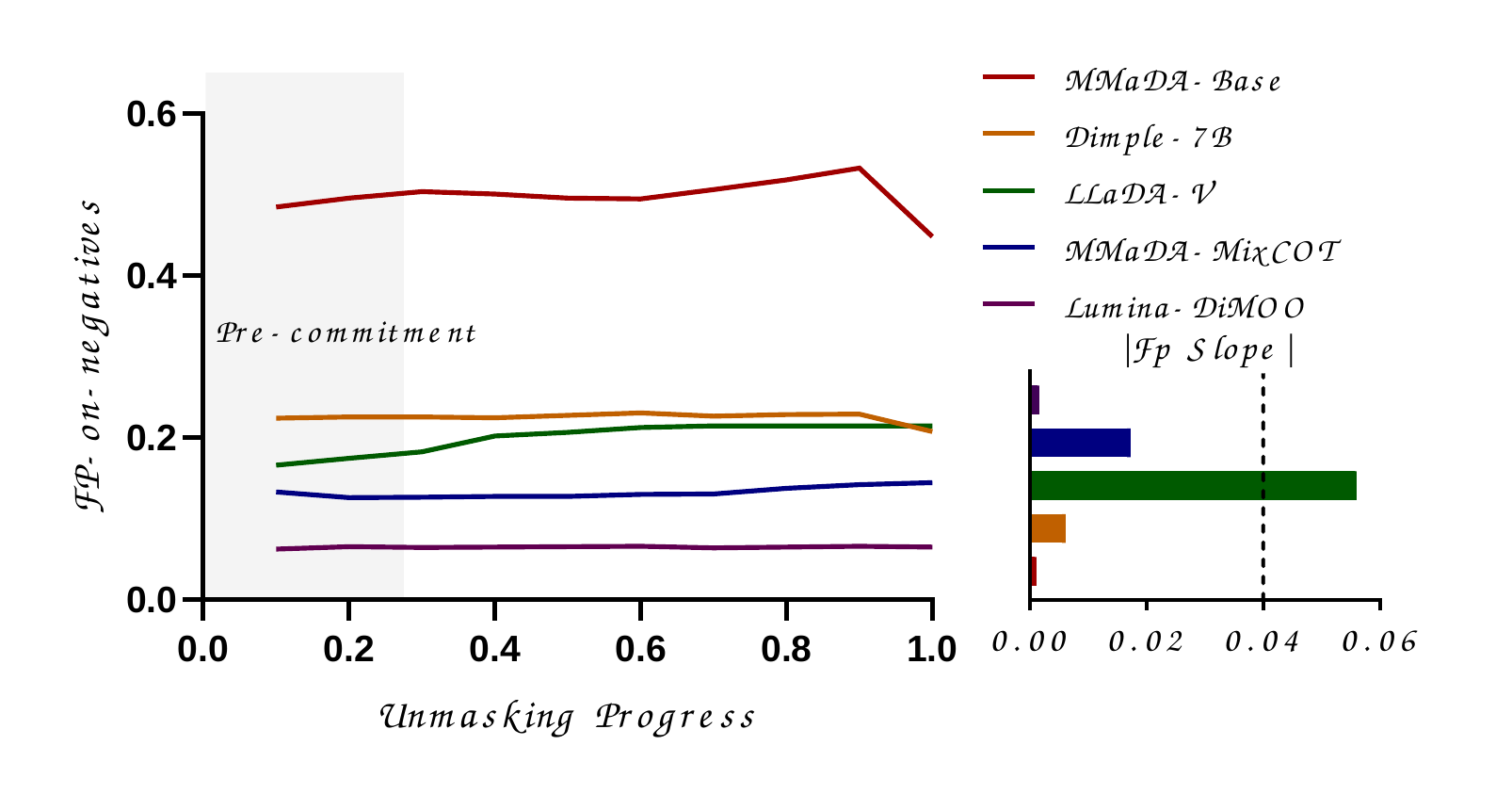}
\caption{Stage-wise false-positive rate on negative propositions across unmasking for five diffusion VLMs (\bench{} test). Rates are read from the tendency state (logit preference at the masked locus) and can therefore differ from the committed FP in Table~\ref{tab:dynahall_leaderboard}. Shaded bands are bootstrap $95\%$ confidence intervals; the inset reports each model's FP slope across the unmasking process.}
\label{fig:flatness}
\end{figure}

\paragraph{Pre-commitment dynamics.}  Across all evaluated models, the pre-commitment prediction agrees with the committed answer on $92$--$98\%$ of trajectories with valid final answers. Figure~\ref{fig:flatness} shows that the stage-wise false-positive curve stays nearly flat throughout unmasking, differing mainly in overall height across models, from roughly $50\%$ for MMaDA-Base to $6.5\%$ for Lumina-DiMOO. Incorrect answers emerge early (normalized emergence $0.06$--$0.15$) and rarely reverse. This pattern is not caused by aggregating correct and incorrect negatives: among final false positives only, the unsupported ``yes'' already dominates the model tendency while the answer position is $80\%$ masked in $77$--$97\%$ of cases as shown by the Early-FP column in Table~\ref{tab:dynahall_leaderboard}. The conversational LLaDA-V shows slightly later emergence, consistent with its post-hoc answer locus. The timing of hallucination therefore remains consistent across architectures, suggesting a general property of masked-diffusion generation. The same early fixation holds for correct answers (early-to-final agreement $92$--$98\%$ per model), so timing alone is not an error signature; the readout instead localizes where the answer is decided, and its statistics still yield a label-free risk score: mean tendency margin over the trajectory predicts committed errors with AUROC $0.60$--$0.78$ across the five models.

\paragraph{Correction, drift, and persistence.} Final-output evaluation conflates three trajectories that \bench{} tracks separately (Figure~\ref{fig:teaser}): an error that persists until commitment, an error corrected before completion, and a correct initial tendency that later drifts into an incorrect commitment. The trajectory-dynamics metrics in Table~\ref{tab:dynahall_leaderboard} show that all three patterns occur, but corrected and drifted trajectories are relatively rare. Across models, $9$--$28\%$ of ever-wrong trajectories recover, and $4$--$12\%$ of ever-correct trajectories drift. On average, a trajectory shows only $0.2$--$1.1$ yes/no flips, with early error persistence as the dominant pattern. These dynamics separate models with nearly identical final accuracy: MMaDA-MixCOT and Lumina-DiMOO differ by only $0.5$ accuracy points but by $4.3\times$ in flips ($0.99$
  vs.\ $0.23$) and $2.3\times$ in correction ($19.9\%$ vs.\ $8.6\%$).

\paragraph{Robustness.} The pre-commitment pattern is not an artifact of the evaluation protocol or probing format. It remains under native parallel decoding: with a $32$-step budget, early-to-final agreement reaches $95$--$97\%$, and the false-positive slope is nearly flat, $[0.00, 0.07]$ across models as shown in the inset of Figure~\ref{fig:flatness}. For the conversational LLaDA-V, enforcing a single-word format, which fixes the answer locus, only strengthens the pattern, increasing early-to-final agreement from $92.9\%$ to $98.5\%$ and reducing invalid outputs from $7.3\%$ to $0.3\%$. Nor is it explained by a yes/no bias: in a four-way multiple-choice control with chance accuracy of $25\%$, the committed option already matches the first-step preference in $88\%$ of MMaDA-Base cases and $95\%$ of Dimple-7B cases. The pattern is also not limited to single-token answers. In open-ended COCO captioning, a hallucinated object is the stable preferred token before it is written in $77.7\%$ $[72.3, 82.6]$ of mentions, leading commitment by a median of two unmasking steps.

\subsection{Structured Failure Modes}
\begin{figure*}[t]
\centering
\includegraphics[width=0.66\textwidth]{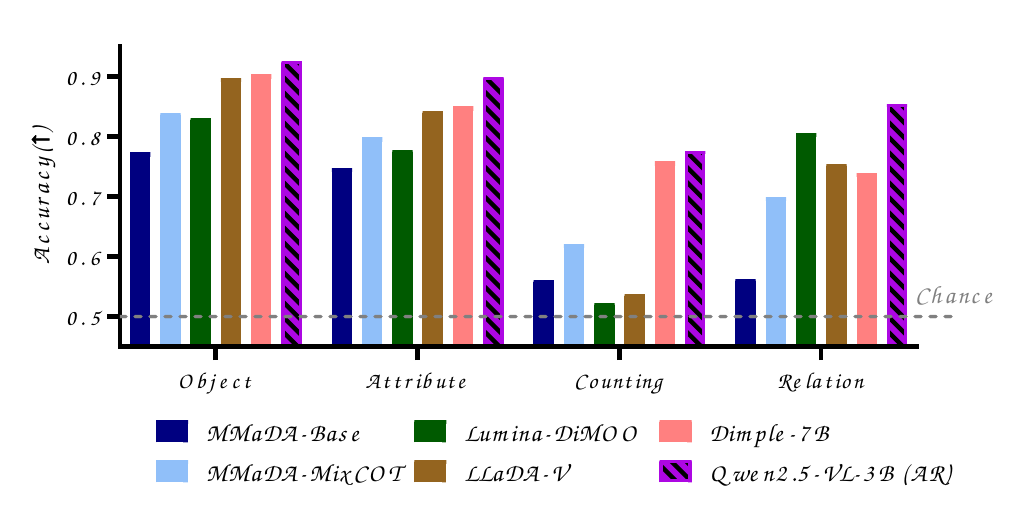}
\caption{Per-task accuracy on \bench{} test ($7{,}000$) for five diffusion VLMs and a strong autoregressive reference (Qwen2.5-VL-3B, hatched); the dashed line marks chance ($0.5$).}
\label{fig:pertask}
\end{figure*}

\paragraph{Task-level collapse.} Analyzing performance by task (Figure~\ref{fig:pertask}; full results in the supplementary material) reveals that object existence is relatively straightforward, with accuracy reaching up to $90\%$. However, performance drops severely on counting and relation tasks across all models. For instance, MMaDA-Base drops to $56\%$ accuracy, with high false-positive rates of $64\%$ on counting and $62\%$ on relations. This collapse is universal but manifests differently: Lumina-DiMOO drives counting false positives to a degenerate $0.2\%$ by rejecting nearly all positive counts (FN $95.4\%$), whereas Dimple-7B, the strongest model overall, still exhibits substantial hallucinations on relations. These specific failure modes are typically obscured by aggregate accuracy metrics or benchmarks focused solely on object existence, such as the widely used POPE.

\paragraph{Generality across generation paradigms.} This performance collapse is linked to the task complexity rather than the decoding mechanism. A strong autoregressive VLM, Qwen2.5-VL-3B~\cite{DBLP:journals/corr/abs-2502-13923}, is the most accurate model overall yet shows the identical task profile as indicated by the hatched bars in Figure~\ref{fig:pertask}: near-ceiling on object existence and attributes but a sharp drop on counting and relations, where it still false-affirms at rates comparable to the diffusion models. What distinguishes diffusion VLMs, and what this benchmark specifically measures, is the exact point along the unmasking trajectory where these errors are formalized.

\paragraph{Prior-driven negatives.} In the object existence task, false positives increase with the strength of language priors. For MMaDA-Base they rise monotonically from $11.5\%$ (random absent) to $30.3\%$ (co-occurring) to $35.2\%$ (semantic-neighbor). Across the other models the prior-laden negatives (co-occurring and semantic-neighbor) remain far harder than random absent objects, even though which of the two is hardest varies as detailed in the supplementary material. Models thus affirm absent objects more readily when priors suggest their presence, confirming that our graded negatives probe visual grounding rather than superficial prior reliance. That these false positives scale with prior strength indicates the early tendency is partly a language-prior readout rather than a completed visual judgment; this prior-driven component is what \method{} targets. Attribute errors are likewise bidirectional by subtype (Table~\ref{tab:attr_subtypes}): shape queries push models toward over-affirmation, material queries toward conservative denial, so aggregating attribute types or testing only color hides both opposing error directions.

\begin{table}[t]
\centering
\footnotesize
\setlength{\tabcolsep}{7pt}

\begin{tabular*}{\columnwidth}{@{\extracolsep{\fill}}llrrr@{}}
\toprule
Model & Attr. & Acc$\uparrow$ & FP$\downarrow$ & FN$\downarrow$ \\
\midrule
\multirow{3}{*}{MMaDA-Base}
 & color    & 72.2 & 20.1 & 35.2 \\
 & material & 65.0 &  9.0 & 61.0 \\
 & shape    & 60.3 & 30.5 & 49.0 \\
\addlinespace
\multirow{3}{*}{Dimple-7B}
 & color    & 85.8 &  7.8 & 20.2 \\
 & material & 86.5 &  2.0 & 25.0 \\
 & shape    & 77.5 & 14.5 & 30.5 \\
\bottomrule
\end{tabular*}
\caption{Attribute-subtype diagnosis for two representative models, on \bench{} test. Columns are accuracy, FP on negatives, and FN on positives (percentages).}
\label{tab:attr_subtypes}
\end{table}

\section{Intervening in the Pre-commitment Window}

The diagnostic value of \bench{} does not depend on a particular mitigation method. Still, the trajectory results suggest a natural hypothesis: if many final false positives already favor the unsupported answer while the locus $p$ is still masked, this pre-commitment window may provide a useful intervention point. We test this hypothesis with \method{} (Pre-commitment Gradient Steering), a lightweight intervention. Let $a_{l,h}$ denote the pre-output-projection activation of head $(l,h)$ at $p$, and let $M=z_{\text{no}}-z_{\text{yes}}$ denote the factual margin, where each $z$ is a log-sum-exp over negative or affirmative surface forms. A hallucinated affirmation therefore gives $M<0$. Since the margin gradient gives the first-order effect of each head on $M$, we score heads by steerability, $c(l,h)=\langle \hat s_{l,h},\bar g_{l,h}\rangle$, where a reference contrast direction $\hat s_{l,h}$ is aligned with the averaged margin gradient $\bar g_{l,h}$. We keep the most steerable heads, ranking by steerability rather than label predictivity, because an accurate predictor need not be an effective intervention point.

To avoid pushing the model toward a generic `no', we build a grounding-oriented direction from hallucinated affirmations $H$ and grounded affirmations $G$, with mean per-head gradients $\bar g_H$ and $\bar g_G$:
\begin{equation}
d_{l,h}=\mathrm{unit}\!\big(\bar g_H-\mathrm{proj}_{\bar g_G}\bar g_H\big).
\end{equation}
This direction removes the component shared with grounded affirmations. We steer only heads that are steerable ($c>0$) and selective between $H$ and $G$, adding $\lambda d_{l,h}$ at $p$ while the locus is masked and stopping once the token is written. Algorithm~\ref{alg:gcs} summarizes the full offline localization and inference-time steering procedure.

\begin{algorithm}[t]
\caption{Pre-commitment Gradient Steering (\method{})}
\label{alg:gcs}
\small
\begin{algorithmic}[1]
\REQUIRE diffusion VLM $\mathcal{M}$; localization set $\mathcal{L}$; affirmation sets $H,G$; \#heads $k$; scale $\lambda$
\ENSURE steered answer at the locus $p$
\STATE Offline: build the steering direction
\STATE accumulate margin gradients $\bar g_{l,h}=\mathbb{E}_{\mathcal{L}}\!\left[\partial M/\partial a_{l,h}\right]$
\STATE score heads $c(l,h)=\langle \hat s_{l,h},\bar g_{l,h}\rangle$; keep the top-$k$ steerable, selective heads $\mathcal{S}$
\STATE $d_{l,h}\leftarrow \mathrm{unit}\!\big(\bar g_H-\mathrm{proj}_{\bar g_G}\bar g_H\big)$ for $(l,h)\in\mathcal{S}$
\STATE Inference: steer inside the pre-commitment window
\FOR{each unmasking step}
  \IF{locus $p$ is masked}
    \STATE $a_{l,h}\leftarrow a_{l,h}+\lambda\, d_{l,h}$ for $(l,h)\in\mathcal{S}$
  \ENDIF
  \STATE apply the native unmasking update of the model
\ENDFOR
\end{algorithmic}
\end{algorithm}

We develop \method{} on MMaDA-8B-Base, the most hallucination-prone model in our study, and verify it on Dimple-7B. We use a stratified $1{,}000$-example test subset for comparisons that require activation access. The baselines are autoregressive mitigations, namely contrastive decoding and activation steering, adapted to operate at the same answer position. On \bench{} (Figure~\ref{fig:main}; full results in the supplementary material), \method{} reduces false positives from $42.7\%$ to $28.5\%$ and improves accuracy by $\ourAccGain$ points, shifting the affirmation rate from $60\%$ toward a balanced $50\%$. Both gains are significant under a paired bootstrap, with $95\%$ CIs of $[-17.2,-11.2]$ and $[+0.8,+5.1]$ points, excluding zero.

Contrastive decoding moves performance in the opposite direction. VCD and DoLa increase false positives, because amplifying the image-conditional distribution worsens the tendency of a model that already over-affirms. Moreover, a diffusion-native margin re-decision strategy, a decision-rule control matched to the false-positive rate of \method{}, produces more false negatives, and a plain final-threshold shift matched to the same $28.5\%$ false positives costs $35.4\%$ false negatives versus $31.1\%$ for \method{}. Because \method{} lowers false positives (from $42.7\%$ to $28.5\%$) at a smaller false-negative cost (to $31.1\%$) than either matched control, it improves the operating frontier rather than merely sliding along it, indicating that the reduction reflects better grounding, not blanket denial. The supplementary material provides further analyses, including head-selection ablations, an image-perturbation control showing that the effect is visually gated, cross-architecture transfer to Dimple-7B, capability checks on POPE, MME, and CHAIR, together with representative qualitative case-study trajectories.

\begin{figure}[t]
\centering
\includegraphics[width=0.92\columnwidth]{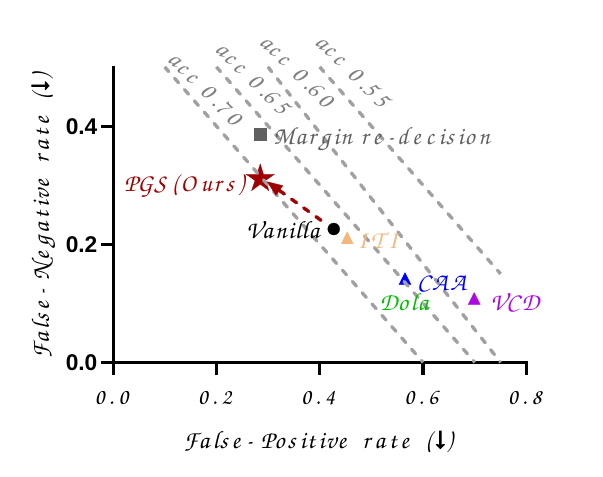}
\caption{Operating points on \bench{}; \method{} is marked by the star. Axes show false-positive and false-negative rates; dashed lines denote iso-accuracy contours.}
\label{fig:main}
\end{figure}

\section{Related Work}
\subsection{Visual Hallucination in Autoregressive LVLMs}
Visual hallucination in large vision-language models (LVLMs) has been studied through benchmarks and mitigations. Benchmarks span caption-level object hallucination~\citep{DBLP:conf/emnlp/RohrbachHBDS18} and question-based probes: polling-based object existence with co-occurrence bias~\citep{DBLP:conf/emnlp/LiDZWZW23}, yes/no perception--cognition items~\citep{DBLP:journals/corr/abs-2306-13394}, existence/attribute/relation coverage without an LLM judge~\citep{DBLP:journals/corr/abs-2311-07397}, and expert-designed language-versus-illusion cases~\citep{DBLP:conf/cvpr/GuanLWXLL0CHYM024}. Mitigations include visual contrastive decoding~\citep{DBLP:conf/cvpr/LengZCLLMB24}, attention penalties and retrospection~\citep{DBLP:conf/cvpr/HuangDZ0H0L0Y24}, post-hoc correction or verification~\citep{DBLP:conf/iclr/ZhouCYZDFBY24,DBLP:journals/corr/abs-2310-16045}, and causal or representation-level edits~\citep{DBLP:conf/iclr/YangL0X25}. All target final faithfulness in autoregressive, final-output evaluation, rather than the intermediate unmasking dynamics that multimodal diffusion models expose.

\subsection{Trajectory Dynamics in Diffusion Language Models}
Diffusion language models generate text by iteratively unmasking a masked sequence under bidirectional context~\citep{DBLP:journals/corr/abs-2502-09992}. This paradigm has recently been extended to multimodal settings, with LLaDA-V, MMaDA, LaViDa, and Lumina-DiMOO building diffusion VLMs~\citep{DBLP:journals/corr/abs-2505-16933,DBLP:journals/corr/abs-2505-15809,DBLP:journals/corr/abs-2505-16839,DBLP:journals/corr/abs-2510-06308}. Because generation unfolds over multiple steps, intermediate trajectories become observable. Recent text-only studies show that hallucination signals can emerge, shift, or self-correct during denoising~\citep{DBLP:journals/corr/abs-2510-01274,DBLP:journals/corr/abs-2602-08048,DBLP:journals/corr/abs-2603-16459,DBLP:journals/corr/abs-2604-26139}. Most closely related to our diagnosis, \citet{DBLP:journals/corr/abs-2508-19982} observe early answer convergence in text diffusion LMs and use it for faster decoding. We instead ask where, along the visual unmasking trajectory, an unsupported answer becomes fixed, and whether later steps ever correct it. For intervention, activation- and distribution-level steering methods have been studied for text diffusion LMs~\citep{DBLP:journals/corr/abs-2512-24143,DBLP:journals/corr/abs-2605-29626,DBLP:journals/corr/abs-2605-10971}, and the concurrent VISAGE~\citep{DBLP:journals/corr/abs-2603-25711} mitigates hallucination in multimodal diffusion models at decoding time by re-ranking token commitments. \method{} instead edits still-masked answer activations within the pre-commitment window and is visually gated. Prior work therefore either focuses on text-only trajectories or studies a specific mitigation, while multimodal diffusion VLMs are still largely evaluated by final answers.

\section{Conclusion}
We have argued that visual hallucination in diffusion VLMs is better understood along the unmasking trajectory than from the final answer alone. \bench{} makes this trajectory measurable. Across five models from three architecture families, hallucination is largely fixed in the pre-commitment window before it is written, a pattern robust to changes in decoding, answer format, and open-ended generation, and structured by task, prior, and attribute type in ways that final-output benchmarks miss. \method{} shows that this window can support intervention, improving the decoding operating point by reducing false positives without degrading general ability. The protocol reads per-step logits and thus targets the growing family of white-box diffusion VLMs; within this setting, we hope \bench{} encourages future work to evaluate and mitigate hallucination where it forms, along the generation trajectory rather than only at the final answer.

\bibliography{aaai2027}

\end{document}